# Unsupervised Decision Forest for Data Clustering and Density Estimation

Hayder Albehadili and Naz Islam

Hmafnd@mail.missouri.edu and IslamN@missouri.edu

**Abstract**

An algorithm to improve performance parameter for unsupervised decision forest clustering and density estimation is presented. Specifically, a dual assignment parameter is introduced as a density estimator by combining Random Forest and Gaussian Mixture Model. The Random Forest method has been specifically applied to construct a robust affinity graph that provides information on the underlying structure of data objects used in clustering. The proposed algorithm differs from the commonly used spectral clustering methods where the computed distance metric is used to find similarities between data points. Experiments were conducted using five datasets. A comparison with six other state-of-the-art methods shows that our model is superior to existing approaches. Efficiency of the proposed model is in capturing the underlying structure for a given set of data points. The proposed method is also robust, and can discriminate between the complex features of data points among different clusters.

## 1. Introduction

Clustering can be described as gathering similar objects, observations, data items, and feature into groups. A number of methods have been used in different contexts for data clustering and the most common method used is known as spectral clustering. Videos, images and many such applications use spectral data clustering to discriminate between different objects. However, it is not an easy task to realize acceptable performance parameters for data clustering by applying spectral clustering because some datasets have very high dimensionality and it is difficult to group similar objects using Euclidean Distance metric. The method itself depends on the accuracy of constructing the affinity matrix, which, to a large degree, depends on the interpretation of the pairwise similarity between clusters. Similarity matrix can also be constructed through such methods as eigenvectors driven from similarity matrix [1, 2, 3]. The accuracy of clustering matrix depends to a large extent on data dimensions.

When the data dimension is very large, it is not an easy task it is construct an affinity graphs or using any simple measurement, e. g. matrix Euclidian distance, because the high dimension observations are usually accompanied by noise and redundant features. In addition, it is difficult to infer density estimation model using Gaussian Mixture Model (GMM) alone for high dimensional data [4]. Finally, it becomes a more difficult undertaking for confining unsupervised task. The contributions of this research are summarized below:

- We use different methods for constructing and evaluating similarity matrix drawn from the random forest.

- Introduction of different algorithm for constructing similarity matrix or affinity graphs. Incorporation of unified affinity graphs and GMM, tuned by Expectation Maximization (EM) algorithm for density estimation model.

- The Random Forest (RF) and GMM have been combined for the first time to produce very robust density estimation which is different from what has been reported recently [5].

## 2. Related work

Data clustering has been widely used for numerous applications [6, 7, 8, 9]. Several methods are used for data clustering. One of the most widely used methods is a spectral clustering as demonstrated in [10, 11, 12, 13]. To overcome spectral clustering issues, Xi Li, et al. [10] proposed a smart method called context-aware hypergraph similarity measure (CAHSM). Also, spectral clustering has been incorporated with Dual Assignment K-Means (SDAKM) by Simon Jones et al. [14]. In addition, several clustering methods were proposed by [15, 16, 17] to enhance spectral clustering, and they have also shown that the relation between pairwise similarities can be established by incorporating different parameters within suggested algorithms. Although using previous suggested methods, spectral clustering is still not adequate to be used with complex datasets. Recently, Xiatian Zhu, et. al. [18] used a random forest to construct affinity graphs. The authors have proposed two subtle methods to construct robust affinity matrix and enhance clustering results. In their implementation, different membership values for partial overlapping input pairwise elements was assigned such that two input elements are assigned a similarity value depending on the path each share. The longer path, the higher is the membership value.

## 3. Clustering using random forest

Random forest is a plurality of trees. Each tree has nodes hierarchically arranged. The information passes from the top of the pyramid continuing to the bottom. Series of inspections from root to the leaves are performed on the transverse patterns. Multitude tests at each node of the tree deterministically isolates irregular or/and dissimilar patterns at each split node. Testing criteria at each weak learner depend on the tasks and especially whether we have supervised or unsupervised learning problem. In this work, since our task is data clustering and density estimation which are unsupervised tasks, concentration will be intensified on unsupervised methods to discriminate between input patterns.

The assumption of input pattern is a feature vector $\chi = (x_1, x_2, \ldots x_d) \in \mathcal{R}^d$. Each node has weak learner which partitions forthcoming patterns according to the following function [5]:

$$h(v, \theta_j) = \mathcal{R}^d \times \mathcal{T} \to \{0,1\} \qquad 1$$

where $\theta_j \in \mathcal{T}$ represents the parameters accompanied with each tree node, $j$ is the jth node, $\mathcal{T}$ is the space for the split parameters, and $v$ is the incoming patterns. We endeavor ultimately to partition arriving data points at each node. It can be tackled using the following:

$$\theta_j = \underset{\theta \in \mathcal{T}}{\arg\max}\; I(\mathcal{S}_j, \theta) \qquad 2$$

It is worth to maximize I in (2) to get high information gain by splitting samples points reaching the node as higher as possible. The split function can be trained using greedy search technique [5]. There are different method to maximize (2) either using Gini or entropy. The entropy can be describe as

$$I = H(\mathcal{S}) - \sum_{i \in \{L,R\}} \frac{|\mathcal{S}^i|}{|\mathcal{S}|} H(\mathcal{S}^i) \qquad 3$$

It is obvious that the higher the information gain is the better splitting node is. The optimization objective function induces partitioning arriving to either left or right channel of the node. Therefore, we consider that the weak learner is the cardinality of the decision forest. Constructing consolidate split functions for decision forest can formulate efficient clustering trees. Since unsupervised learning is demanded, following [5], below is the partitioning paradigm formulation used at each weak learner

$$I(S_j, \theta) = H(S_j) - \sum_{i \in \{L,R\}} \frac{|S_j^i|}{|S_j|} H(S_j^i) \qquad 4$$

Then the entropy can be defined as

$$H(S) = \frac{1}{2} \log((2\pi e)^d |\Lambda(s)|)$$

Then the information gain can be obtained as following

$$I(S_j, \theta) = \log(|\Lambda(S_j)|) - \sum_{i \in \{L,R\}} \frac{|S_j^i|}{|S_j|} \log(|\Lambda(S_j^i)|) \qquad 5$$

Where $\Lambda$ is dxd covariance matrix and $|.|$ is a determinant for the matrix.

### 4. A brief review contemporary existing methods

A contemporary approach proposed framework [18] is utilized the partial overlapping and they give different degree of memberships for different length of partial overlapping, which is the intrinsic difference with binary model. There are two paradigms were demonstrated. The affinity matrix for the first model is inferred below

$$a_{ij}^t = \frac{\ell}{\max(|\rho^i||\rho^j|) - 1} \qquad 6$$

Where $\ell$ is the length of partial lapping between pairwise element and $\rho^i$ and $\rho^j$ are path lengths for both samples $\chi_i$ and $\chi_j$ respectively. In this framework, the depth of tree nodes is equally weighted, means that the weights are equally distributed. This model called ClustRF-Strct-Unfm. To mitigate limitations in the first model, the supplementary model is also proposed by [18] called Adaptive Structure Model considered better than the ClustRF-Strct-Unfm. It is obvious that this work concentrate on how constructing affinity graph by mainly depending on how pairwise elements shared paths along decision trees. However, the most important issues that should be considered are how the incoming patterns discriminate? What kind the weak learners have been used inside trees? Are those split functions used in decision forest sufficient to split high dimensional data? All these questions are considered in our implementation in next sections.

### 5. The proposed decision forest

To address confines in [18] and capture the intrinsic underlying data semantic, instead linear and nonlinear model are incorporated in our decision forest.

adequate to retrieve robust affinity matrix used for spectral clustering later.

To address confines in [18] and capture the intrinsic underlying data semantic, instead linear and nonlinear model are incorporated in our decision forest.

$$h(v, \theta) = [\tau_1 > \phi(v). \psi > \tau_2] \qquad 7$$
$$h(v, \theta) = [\tau_1 > \phi^T(v) \psi \phi(v) > \tau_2] \qquad 8$$

Moreover, to identify more complex relation between input patterns, we also propose GMM to be incorporated to the random forest because it is very robust method for unsupervised data clustering [19, 20]. In out implementation, we use each weak learner at a certain depth of random forest. The reason of using several weak learners is because diversity can lead to more generalization for capturing different data associations. To beset of our knowledge, this is the first time decision forest trained with several and complex weak learners. For constructing affinity matrix we use the same what have been proposed in [18].

### 5.1 Dual Assignment to construct affinity matrix

In all models above, the notion is still assigned membership values for all incoming samples in all circumstances whether the samples are partially or entirely overlapped. In this part, imposing those only patterns that have the smallest similarity membership will be refrained and the patterns having highest similarity will be induced. We qualify the samples by adjusting a threshold. According to eqn. (8), an input pattern $(\chi_i, \chi_j)$ can have a maxim similarity in defining affinity matrix if they are survived along a path together. However, they still also have some degree of similarities according to a used model even they are entirely from different clusters. Furthermore, imagine that we have a pair of patterns $\chi_i$ and $\chi_j$ and they belong to the same cluster, then they will be assigned minimum degree of similarity in the constructed affinity matrix. Thus, this assumption is not true especially for complex irregular patterns, which cannot perfectly disassemble the similar examples related to the same clusters. This is the nature of the weak learners because they are only thresholds capturing inequalities trying to separate correlating samples affined to one cluster from other samples united to different clusters. Intuitively, since split function always not perfectly dissociate clusters, otherwise, we would end with random forest with only just few nodes, then random forest will incorrectly cluster associated data points. For example, let assume we have six input samples $(\chi_1, \chi_2, ..\chi_6)$ represented by three clusters $\{(\chi_1, \chi_2), (\chi_1, \chi_2), (\chi_1, \chi_2)\}$ as shown in fig. 1. Obviously the affinity matrix is a constituent element of visually estimating the number of clusters. For ideal clustering, the number of clusters around the diagonal affinity matrix must be three clusters and having a membership one for data points occurring within the same cluster and zeros for the others. However, in all clustering methods, we still have a membership assigned to other clusters as represented in x in fig. 1 above. Furthermore, the correlating samples nested in the diagonal matrix have membership less than one and it could be less than the similarity between pairwise elements not occurring around the diagonal. They can be seen clearly in the similarity graph, which shows that there are three clusters as well as some data points that are ambiguous to belong to one of the three clusters.

Obviously, it can be seen that random forest is very confident to capture similarity between points, which have values equal ones. However, there are more values which make difficult to random forest to judge to which relative cluster is belong to. In our model, we use a threshold $ʒ$ as a filter to extract input patterns having strong relation from other patterns and set patterns with low relation to be having zeros. Then pass sample points which have values bigger that threshold $ʒ$ to GMM. Not like contemporary GMM which has no prior information about latent variables. In our mode, GMM is informed about how data clusters belong to same clusters look like because some of the hidden variables are devoted from the decision forest. In this case, GMM has higher chance to assign vague samples to affined clusters. Furthermore, since some of hidden values previously known, GMM requires less time training and more confident finding the related clutters for those samples points. Because this assumption, a constraint will be used to eliminate pair data points having low similarity measure.

Example:- in this example a dataset that has three overlapping clusters have been created with 800 (2D for easy visualization) samples drawn in fig. 2 (a). The first cluster has 175 data points, second one has 250 samples, and the third one has 375 samples. In this example, only the random forest proposed in [18] will be compared to our model for abbreviation and both of them use random forest to construct the affinity matrix. The qualitative

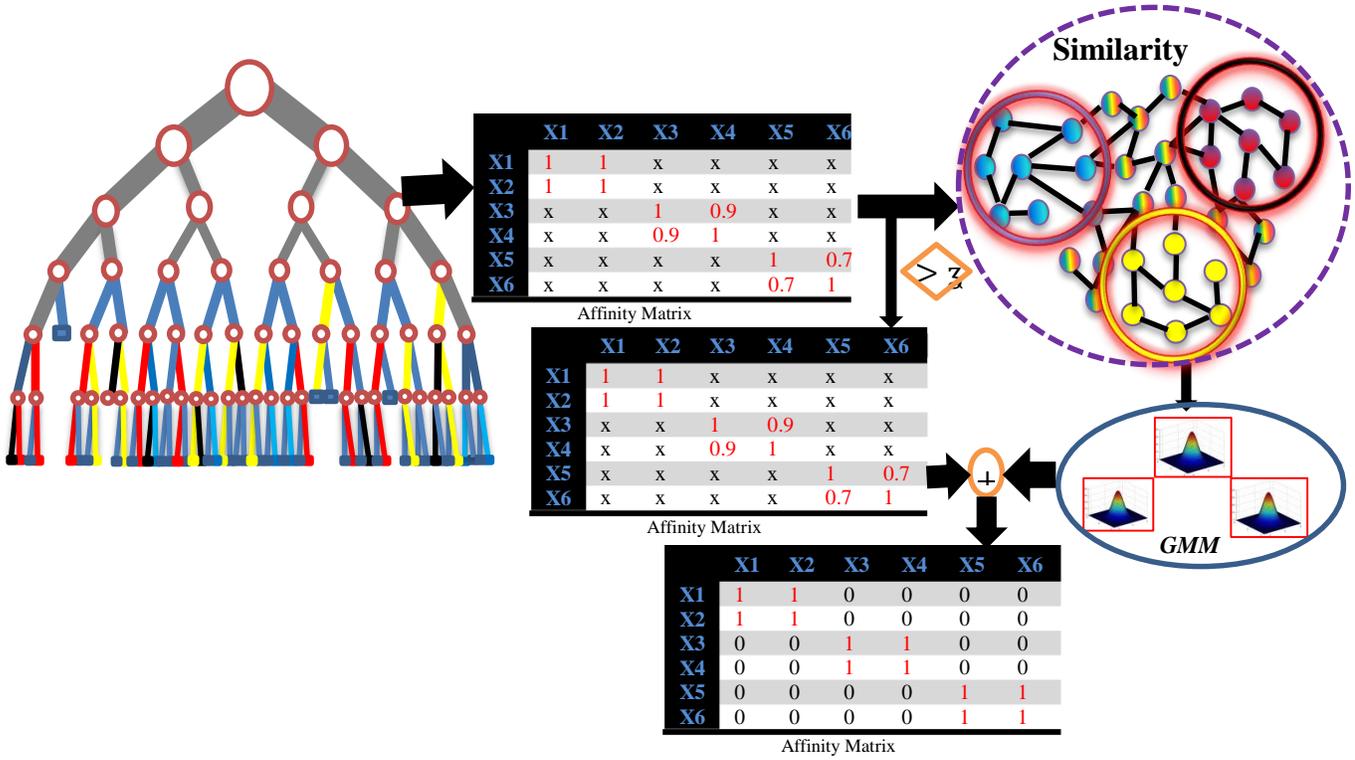
Fig.1. clustering six input samples using random forest and GMM

clustering using binary model, ClustRF-Strct-Adpt, and ClustRF-Strct-Unfm [18] is shown in fig. 2 (b), (c), and (d) respectively. It is clear that the blocks of affinity matrices drawn from previous model are not distinct although experiment is conducted with plurality of trees (1000 tress). Nevertheless, the DRFGMM model has more evident blocks than the other models because the noise has been diminished. we quantitatively compare with other methods as shown in table 1. The accuracies in this example are 29.6%, 29.8%, 40.8% and 89.5% representing binary, ClustRF-Strct-Unfm, ClustRF-Strct-Adpt, and DRFGMM models respectively.

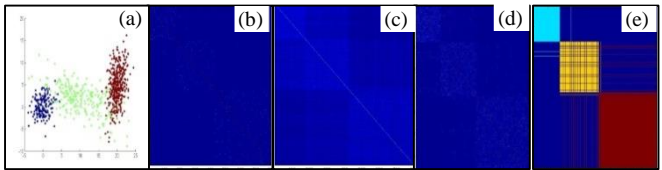
Fig.2. Results of using different random forests

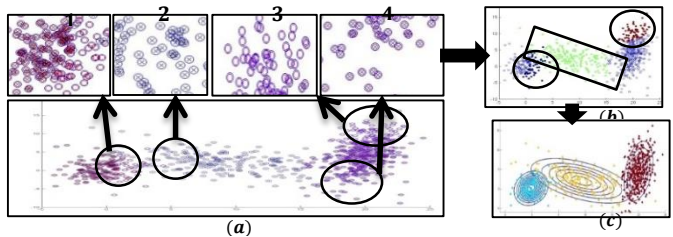
Fig. 3. Example of dual processing using RF and GMM (a) Affinity matrix after threshold (b) Affinity matrix after mutually exclusive

Fig. 3 (a) shows the results after slicing affinity matrix, for clarifications, and some regions inside the figure are zoomed outa and numbered. The blue circles represent the original data points. The x marks represent one of the three clusters, red circles on the right and left sides represent the other clusters. Some data points as showed in the first upper left are shared by more than one cluster and also it is clear in the upper left (box 4) that the box some of data points shared by other clusters. However, in the both box 1 and 2, those clusters are mutually exclusive. Afterwards, the points that have been shared by more than one cluster are excluded as shown in fig. 3 (b). The remaining points are marked with circles and rectangle as well. Subsequently, those points are passed into GMM for final clustering and the results showed in fig. 3 (c). The block diagram showed in Fig. 4 exhibits the dual processing stages of RF and GMM.

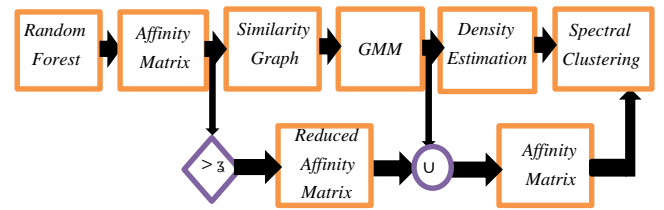
Fig. 4. Dual processing using RF and GMM presentation

### 5.2 Updating equations for GMM and RF (GMM-RF)

In this part, we delve the relation between random forest and GMM and how we incorporate the two models. The intrinsic formula for GMM is give below

$$\mathcal{P}(x|\theta) = \sum_{j=1}^{m} a_j \, \mathcal{P}(x|z_j, \theta_j)$$

Where x is an observation, $z_j$ is the latent variables, and $\theta_j$ is the associated parameters with GMM. Gaussian normal distribution is used in this implementation. The final distribution is given below

$$\mathcal{P}_j(x_i|\mu_j, \Sigma_j) = \frac{1}{(2\pi)^{1/2}|\Sigma|^{1/2}} \exp\left(-\frac{1}{2}(x_i - \mu_j)\Sigma_j^{-1}(x_i - \mu_j)\right)$$

The latent/hidden variables can be obtained using the following formula

$$\mathcal{Z}_{ji} = \mathcal{P}(\mathcal{Z}_{ji}|x_i, \theta^{old}) = \frac{\varepsilon_j \mathcal{P}(x_i|\theta_j^{old})}{\sum_{k=1}^{m} \varepsilon_k \mathcal{P}(x_i|\theta_k^{old})} \quad (9)$$

To mitigate limitations of GMM because all hidden variables $\mathcal{Z}_{ji}$ are not known, we alleviate the ambiguity inherited by absence all the latent variables, ensembles of these variables are afforded by decision forest. Revival GMM by decision forest deterministically tackles fogginess of conventional GMM. Assuming that the latent variables deriving from random forest embedded within $\mathcal{RF}_{ji}$ therefore the (3.6) can be updated as below

$$\mathcal{Z}_{ji}|(\mathcal{Z}_{ji} \neq \mathcal{RF}_{ji}) = \mathcal{P}(\mathcal{Z}_{ji}|\mathcal{Z}_{ji} \neq \mathcal{RF}_{ji}|x_i, \theta^{old})$$
$$= \frac{\varepsilon_j \mathcal{P}(x_i|\theta_j^{old})}{\sum_{k=1}^{m} \varepsilon_k \mathcal{P}(x_i|\theta_k^{old})} \quad (10)$$

Then the updating equations for parameters of GMM can be update ed as following

$$\varepsilon_j^{new} = \frac{1}{N}\sum_{i=1}^{N}(\mathcal{Z}_{ji} \cup \mathcal{RF}_{ji}) \quad (11)$$

$$\Sigma_j^{new} = \frac{\sum_{i=1}^{N}(\mathcal{Z}_{ji} \cup \mathcal{RF}_{ji})(x_i - \mu_j^{new})(x_i - \mu_j^{new})}{\sum_{i=1}^{N}(\mathcal{Z}_{ji} \cup \mathcal{RF}_{ji})} \quad (12)$$

$$\mu_j^{new} = \sum_{i=1}^{N} \frac{(\mathcal{Z}_{ji} \cup \mathcal{RF}_{ji})x_i}{(\mathcal{Z}_{ji} \cup \mathcal{RF}_{ji})} \quad (13)$$

It is obvious now that GMM knows some of hidden variables given from the random forest.

## 6. Experimental results

### 6.1 Comparison with five State-Of-The-Arts

Our models are compared with six state-of-the-art implementations.

1. Uniform Structure Model (ClustRF-Strct-Unfm) and Adaptive structure-aware affinity inference (ClustRF-Strct-Adpt) [18].

2. Graph regularized Non-negative Matrix Factorization (GNMF)[21]: This work uses a popular method called Non-negative Matrix Factorization (NMF), which aims to factorize non-negative data matrix into non-negative data matrices approximated to original matrix.

3. Sparse Concept Coding (SCC) [22]: It is an accomplished method for matrix factorization. they this use Sparse Concept Coding (SCC) algorithm, which consists of two stages representing learned basis step and representation step.

4. Locally Consistent Gaussian mixture Model (LCGMM) [23]: To alleviate GMM limitations, they introduce a GMM which is unified with a regularizer which induces geometry of marginal distribution by considering into account elevation of nearest neighbor graph.

5. Landmark-based Spectral Clustering (LSC) [24]: This method has two main parts to select Landmark selection either random selection or using k-means then they called LSC-R and LSC-K respectively.

6. Normalized Cuts (Ncut) [25]: it is one of robust clustering algorithms. The notion of Ncut is that the images are partitioned into groups.

### 6.2 Datasets

To evaluate the models, several clustering experiments were conducted using non-trivial ensemble datasets, including one dataset in digits and two datasets each for image and video. The datasets are summarized in Table 1, and samples from three of the datasets are shown in fig. 5 and 6. Two evaluation metrics are used to compute performance accuracy. We use Accuracy (ACC) and Normalized Mutual Information (NMI) [21]. The following are the descriptions for our benchmarks:

(1) USAA [26] datasets: it has eight YouTube videos for different semantic classes. It is an intricate dataset because it has several unconstrained social effectiveness.

(2) CMU-PIE [21]: it is a gray (32x32 pixels) scale face images. In addition, the dataset has 68 persons and each person images have different illuminations and poses.

(3) ERCe [27]: the dataset has several human activities taken from six campus events and involves 600 videos. The dataset is not trivial because physical activities are unconstraint.

(4) COIL20 [21]: this dataset has 20 objects. Each object has 72 (32x32) gray scale images. Various angles are associated with those objects.

(5) The MNIST [28] is a hand written digits 0-9. The dataset consists of 10000 samples. All samples have the same 28x28 pixels size. The pixels are scaled to the [0, 1] before training.

**Table 1: statistics of the datasets used in this experiments**

| Dataset | No. Samples | Dimensionality | No. of clusters |
|---|---|---|---|
| **MNIST** | 10000 | 1024 | 10 |
| **USAA** | 1466 | 14000 | 8 |
| **ERCe** | 600 | 2672 | 6 |
| **COIL20** | 1440 | 1024 | 20 |
| **PIE** | 2856 | 1024 | 68 |

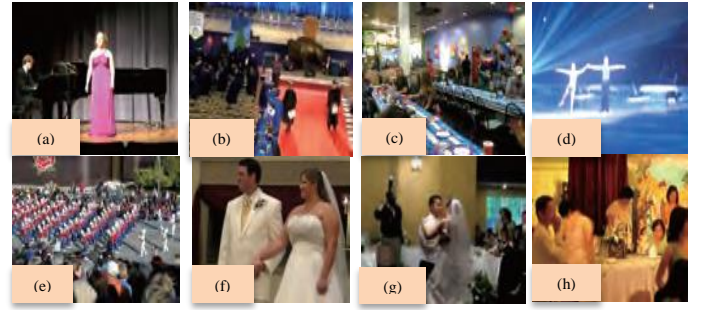

Fig. 5. USAA dataset[26]: (a) Music performance (b) Graduation (c) birthday (d) Non music performance (e) Parade (f) wedding ceremony (g) wedding dance (h) wedding reception

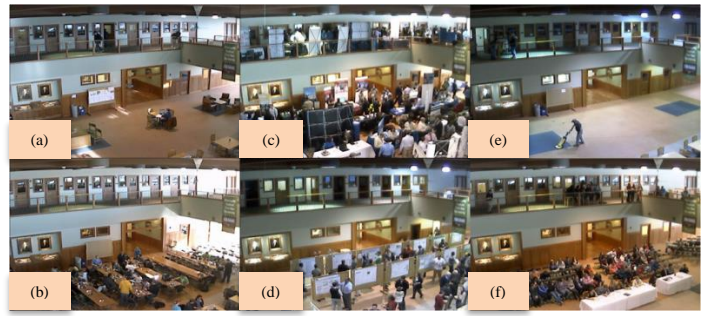

Fig. 6. ERCe dataset[27]: (a) Student Orientation (b) Group Studying (c) Career Fair (d) Forum on Gun Control and Gun Violence (e) Scholarship Competition (f) Cleaning.

### 6.3 Implementation and parameters details

In all the experiments the numbers of trees in each mode were chosen to be 200, where the number of try in split function depends on weak learners, and the complexity of the dataset. In this study, the number of linear split functions, $m_{try}$ and the number of trees are set to 5 and 200 respectively as opposed to 5 and 1000 as in [18]. In addition, we compare the efficiency based on the techniques described with those of researchers listed below:

- In [21]: The performance is stable for when the value of λ ranging between 10 and 1000. However, the performance drastically decreases as the p increases. Therefore, both λ and p are set to 100 and 3 respectively.

- Sparse Concept Coding (SCC) [22]: the number of nearest neighbors and regularization parameter are left same as in [21] to be 5 and 0.1 respectively. The number of basis vectors is equally set to the number of clusters in each the dataset.

- Locally Consistent Gaussian mixture Model (LCGMM) [23]: we follow [23] using the same default parameters setting (the number of nearest neighbors p is 20 and λ is 0.1)

- Landmark-based Spectral Clustering (LSC) [24]: following [24], the same default parameter setting are used (the number of landmarks is 500 and the number of nearest landmarks is 6).

- Normalized Cuts (Ncut) [25]: the same default parameters are used.

### 6.4 Clustering Performance

We conducted experiments on five datasets and Table (2) compares our results with six previous methods (eight different approaches). We achieve 4.9% and 16.3% against the first and second best models on the PIE; 0.8% and 2% on USAA, 11.5% and 12.6% on ERCe, 4.3% and 4.8% on COIL20, and 0.8% and 4.1% on MNIST. In addition, superiority is alternative between our RF and DERGMM. We also tested our models through experiments on different randomized numbers of clusters, as was done previously [33, 23, 22]. The results, conducted on datasets COIL20, PIE and MNIST are shown in Tables 3, 4, and 5 respectively.

We also conducted randomized experiments on different number of clusters [21], for every test, we run a random number of clusters are chosen for each given n clusters. The results are shown in tables 3, 4, and 5 for COIL20, PIE, and MNIST datasets respectively. It is obvious that our methods always surpass to all other compared methods. 6.5 Effect of dimensionality reduction

In order characterize our methods with respect to the dimensions of the datasets; experiments were conducted with varying dimensional for each given dataset. Results show that dimensionality reduction can influence the overall performance. Fig. 7 shows the results for all the datasets. In each figure the left side column represents accuracy; the right side column represents the size of the captured covariance accompanied with dimension.

It is worth mentioning that in fig. 7c, which represents the results of MNIST dataset, the results for both ClustRF-Strct-Unfm and ClustRF-Strct-Adpt are not included because both methods run very slow and they require huge amount of memory, which cannot be provided although the experiments run using Dell XPS 8500, core i5, and 8GB memory and it quickly signals memory full after few iteration. The reason behind it is because the dataset is big and the two approaches used 1000 tress and each one can continue branching for unlimited depth.

### 7. Conclusion

In this work, we propose two methods to enhance data clustering. The first model used is the enhanced Random Forest (RF). Three different functions are incorporated into split functions to induce more robust RF than what had been proposed in [18]. Weak learners are included in both linear and nonlinear functions distributed on a certain levels on each tree of the random forest. Furthermore, RF is consolidated by GMM inserted between linear and linear functions. Consequently, robust RF is appropriately able to discriminate between uninformative features because it discovers semantic underlying structure data.

In second proposed method, dual assignment is suggested using both RF and GMM (DRFGMM). This method diminishes laminations from affinity matrix constructed by RF. After affinity matrix is passed into next stage of the GMM, only the pairwise elements that have high similarity membership will be extracted from the matrix and exclude all other members. In addition to consolidate the affinity matrix constructed from RF, density estimation is also is another part of the implementations. Extensive experiments have been conducted on data clustering. Five benchmark datasets are used for the evaluation. Moreover, we compare our model with six robust state-of-the-art methods. We show our models are superior to all other models. To the best of our knowledge, this is the first elaboration between random forest and GMM introduced for data clustering and density estimation.

**Table 2: Clustering accuracy on the five datasets**

| Dataset | | CMU-PIE | USAA | ERCe | COIL20[33,32] | MNIST |
|---|---|---|---|---|---|---|
| GNMF | [21] | 77.4 | 7.4 | 60.5 | 75.0 | 73.0 |
| LSC-R | [24] | 62.7 | 3.7 | 53.6 | 68.4 | 73.5 |
| LSC-K | [24] | 65.9 | 3.4 | 60.0 | 66.7 | 76.3 |
| LCGMM | [23] | 63.4 | 4.3 | 61.6 | 66.1 | 73.6 |
| SCC | [22] | 60.5 | 6.2 | 63.7 | 74.5 | 71.0 |
| Ncut | [25] | 57.7 | 3.2 | 49.1 | 69.6 | 71.8 |
| ClustRF-Strct-Unfm[18] | | 48.2 | 4.7 | 59.3 | 61.4 | - |
| ClustRF-Strct-Adpt[18] | | 52.3 | 5.7 | 60.4 | 71.0 | - |
| DRFGMM | [this work] | 82.3 | 3.6 | 70.9 | 79.3 | 73.5 |
| RF | [this work] | 79.2 | 8.2 | 73.1 | 74.6 | 77.1 |

**Table 3 (a): Clustering performance on COIL20 dataset-Accuracy metric**

| COIL20 Dataset-Accuracy (%) | | | | | | | | | | |
|---|---|---|---|---|---|---|---|---|---|---|
| K | GNMF | LSC-R | LSC-K | LCGMM | SCC | Ncut | Strct-Unfm | Strct-Adpt | DRFGMM | RF |
| 4 | 80.2±25.5 | 74.2±4.1 | 82.2±6.3 | 79.3±8.3 | 89.3±6.3 | 86.2±0.3 | 65.8±25.3 | 87.4±11.2 | 97.7±2.3 | 88.6±1.3 |
| 8 | 83.2±4.4 | 77.5±1.8 | 79.1±4.9 | 76.2±4.6 | 83.4±4.9 | 81.1±0.7 | 57.4±5.9 | 74.6±9.8 | 88.6±1.5 | 83.0±0.8 |
| 12 | 87.2±1.3 | 76.0±1.4 | 71.6±1.8 | 78.2±5.0 | 88.6±1.8 | 76.2±0.8 | 62.2±1.0 | 79.5±12.4 | 83.3±9.0 | 81.2±9.9 |
| 16 | 82.3±5.9 | 71.7±4.7 | 74.4±0.4 | 75.3±2.7 | 80±3.3 | 68.9±9.0 | 72.5±8.8 | 62.7±09.4 | 83.0±6.9 | 79.1±6.4 |
| 20 | 75.0 | 68.4 | 66.7 | 66.1 | 74.5 | 69.6 | 61.4 | 71.0 | 79.3 | 75.6 |
| **Avg.** | 81.5 | 72.4 | 74.8 | 75.0 | 83.1 | 76.4 | 63.86 | 75.0 | **86.3** | 81.5 |

Table 3 (b): Clustering performance on COIL20 dataset-NMI metric

| COIL20 Dataset-Normalized Mutual Information (%) | | | | | | | | | | |
|---|---|---|---|---|---|---|---|---|---|---|
| K | GNMF | LSC-R | LSC-K | LCGMM | SCC | Ncut | Strct-Unfm | Strct-Adpt | DRFGMM | RF |
| 4 | 84.1±15.8 | 76.5±15.0 | 83.4±6.8 | 87.4±3.8 | 94.1±13.1 | 89.1±8.06 | 75.8±19.7 | 92.3±2.1 | 97.7±4.7 | 98.6±2.8 |
| 8 | 88.5±11.0 | 87.8±18.0 | 87.4±10.8 | 85.1±21.0 | 84.8±20.0 | 87.1±14.9 | 69.3±14.3 | 84.0±1.2 | 96.1±4.7 | 94.8±1.9 |
| 12 | 92.7±0.9 | 89.0±1.0 | 81.8±3.054 | 87.4±18.2 | 91.1±1.1 | 81.2±16.1 | 77.1±7.9 | 88.1±8.1 | 99.0±2.7 | 93.9±4.1 |
| 16 | 91.2±5.0 | 82.2±1.1 | 84.9±3.6 | 83.4±20.5 | 86.7±1.9 | 77.5±2.9 | 78.0±8.4 | 75.2±6.0 | 89.8±3.5 | 89.7±3.6 |
| 20 | 85.6 | 79.4 | 81.4 | 79.0 | 84.1 | 76.4 | 74.2 | 82.8 | 91.4 | 88.4 |
| Avg. | 88.4 | 81.36 | 83.78 | 84.4 | 88.1 | 82.2 | 74.8 | 84.4 | **94.8** | 93.0 |

Table 4 (a): Clustering performance on PIE dataset-Accuracy metric

| PIE Dataset- Accuracy (%) | | | | | | | | | | |
|---|---|---|---|---|---|---|---|---|---|---|
| K | GNMF | LSC-R | LSC-K | LCGMM | SCC | Ncut | Strct-Unfm | Strct-Adpt | DRFGMM | RF |
| 10 | 89.1±3.5 | 87.0±8.3 | 88.8±9.0 | 88.2±19.0 | 95.4±1.0 | 95.4±2.7 | 61.1±7.7 | 85.9±.3.5 | 91.7±4.2 | 94.7±1.7 |
| 20 | 81.6±5.5 | 85.0±4.6 | 83.8±19.1 | 79.5±10.8 | 88.9±1.3 | 81.0±4.4 | 71.3±15.9 | 79.2±0.2 | 96.9±2.1 | 84.4±4.1 |
| 30 | 79.8±3.2 | 78.7±5.0 | 78.4±0.11.2 | 76.3±5.4 | 90.9±7.9 | 72.8±1.3 | 77.6±6.6 | 68.1± 3.1 | 88.3±2.3 | 88.8±4.6 |
| 40 | 77.3±3.5 | 73.3±2.7 | 74.4±5.6 | 65.1±2.7 | 72.8±3.9 | 63.0±4.4 | 66.0±03.3 | 60.5±1.9 | 82.6±1.1 | 81.0±3.1 |
| 50 | 74.9±3.8 | 69.9±1.3 | 68.5±9.9 | 64.6±7.3 | 64.6±8.8 | 59.3±5.8 | 62.2±3.0 | 59.6±1.5 | 85.7±5.4 | 81.1±0.5 |
| 60 | 73.3±1.2 | 67.1±0.8 | 67.1±5.0 | 74.6±03.8 | 78.7±.9 | 73.8±2.6 | 58.0±1.4 | 63.0±1.5 | 73.5±2.3 | 79.6±2.1 |
| 68 | 77.4 | 62.7 | 65.9 | 63.4 | 60.5 | 57.7 | 48.2 | 52.3 | 82.3 | 79.2 |
| Ave. | 79.1 | 74.1 | 75.2 | 73.1 | 78.8 | 71.8 | 63.4 | 66.9 | **85.8** | 84.1 |

Table 4 (b): Clustering performance on PIE dataset-NMI metric

| PIE Dataset-Normalized Mutual Information (%) | | | | | | | | | | |
|---|---|---|---|---|---|---|---|---|---|---|
| K | GNMF | LSC-R | LSC-K | LCGMM | SCC | Ncut | Strct-Unfm | Strct-Adpt | DRFGMM | RF |
| 10 | 81.7±2.7 | 91.0±1.3 | 92.0±3.0 | 93.6±6.7 | 95.3±1.5 | 95.3±0.8 | 73.0±2.1 | 93.1±3.1 | 98.5±2.3 | 97.0±3.2 |
| 20 | 89.5±3.2 | 88.9±7.9 | 99.0±4.4 | 88.1±4.4 | 91.5±0.8 | 89.3±0.3 | 79.1±2.4 | 88.7±1.9 | 99.6±2.1 | 92.6±1.5 |
| 30 | 89.9±1.5 | 87.8±3.9 | 80.0±4.8 | 88.6±2.1 | 94.4±5.5 | 82.7±3.8 | 86.8±1.5 | 78.6±4.0 | 94.8±3.8 | 94.2±2.7 |
| 40 | 89.7±1.3 | 81.3±8.8 | 86.2±8.1 | 83.3±0.9 | 82.7±3.2 | 81.9±0.2 | 76.1±1.7 | 68.3±1.8 | 93.9±1.1 | 89.3±1.8 |
| 50 | 89.1±1.1 | 84.1±5.9 | 78.7±5.2 | 61.7±3.5 | 61.7±3.3 | 73.1±0.2 | 77.0±0.6 | 73.8±9.2 | 93.4±1.5 | 91.6±3.5 |
| 60 | 89.9±1.9 | 76.8±8.4 | 77.2±6.7 | 84.0±2.7 | 87.8±2.3 | 73.8±0.9 | 74.1±4.8 | 74.7±6.0 | 85.5±1.2 | 98±1.3 |
| 68 | 88.0 | 75.2 | 77.8 | 76.1 | 68.3 | 75.1 | 59.3 | 64.6 | 88.6 | 89.7 |
| Ave. | 88.2 | 83.5 | 83.2 | 82.2 | 83.1 | 81.6 | 75.0 | 77.4 | **93.4** | 92.1 |

Table 5 (a): Clustering performance on MNIST dataset-Accuracy metric

| MNIST Dataset- Accuracy (%) | | | | | | | | |
|---|---|---|---|---|---|---|---|---|
| K | GNMF | LSC-R | LSC-K | LCGMM | SCC | Ncut | DRFGMM | RF |
| 3 | 96.8±0.2 | 96.9±4.0 | 91.7±7.5 | 91.2±4.9 | 87.3±6.3 | 86.9±4.9 | 91.7±0.06.3 | 91.7± 5.0 |
| 6 | 79.5±0.1 | 79.6±2.0 | 77.2±7.0 | 86.1±3.1 | 78.1±6.3 | 77.6±3.0 | 82.6±0.031.1 | 88.3± 0.4 |
| 10 | 73.0 | 73.5 | 76.3 | 73.6 | 71.0 | 71.8 | 73.5 | 77.1 |
| Ave. | 82.2 | 83.3 | 81.7 | 83.6 | 78.5 | 78.7 | 82.6 | 85.7 |

Table 5 (b): Clustering performance on MNIST dataset-NMI metric

| MNIST Dataset-Normalized Mutual Information (%) | | | | | | | | |
|---|---|---|---|---|---|---|---|---|
| K | GNMF | LSC-R | LSC-K | LCGMM | SCC | Ncut | DRFGMM | RF |
| 3 | 93.3±4.1 | 97.6±1.2 | 94.5±1.1 | 93.3± 1.3 | 93.7± 5.0 | 93.4± 0.4 | 95.0± 6.6 | 96.6± 0.9 |
| 6 | 84.8±6 | 98.0±2.1 | 88.7±3.6 | 93.5± 3.9 | 88.2± 5.3 | 89.1± 0.95 | 93.1± 0.8 | 95.8± 0.4 |
| 10 | 78.7 | 81.5 | 87.6 | 89 | 81.4 | 81.0 | 82.5 | 89.8 |
| Ave. | 85.6 | 92.3 | 92 | 89.2 | 87.7 | 87.8 | 92.0 | **94.0** |

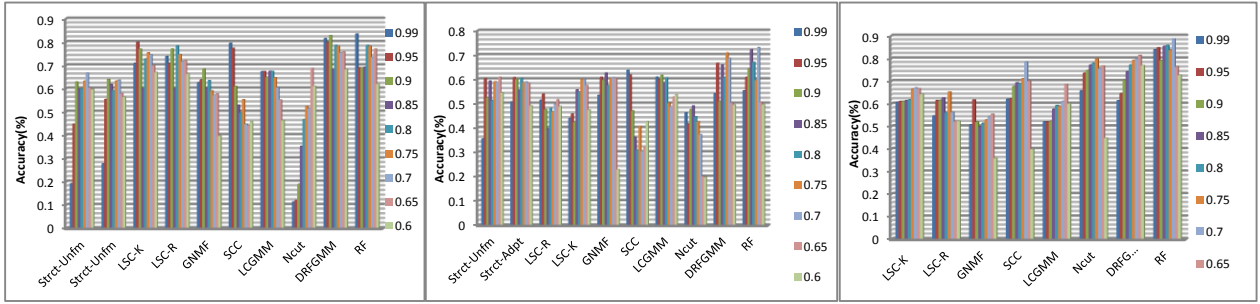

Fig. 7. Accuracy versus dimensionality reduction (a) COIL20 (b) ERCe (c) MNIST datasets.